\documentclass{esannV2}
\usepackage{graphicx,url,hyperref}
\usepackage{xcolor,wrapfig,algorithmic}
\usepackage[scriptsize,tight]{subfigure}
\usepackage{tabularx}
  \newcolumntype{C}[1]{>{\centering\arraybackslash}p{#1}}
  \newcolumntype{R}[1]{>{\raggedleft\arraybackslash}p{#1}}
  \newcolumntype{L}[1]{>{\raggedright\arraybackslash}p{#1}}
\usepackage{booktabs} 
\usepackage[ruled,linesnumbered]{algorithm2e}
\usepackage{amssymb}
\usepackage{amsmath}
\usepackage{mathtools}
\usepackage{xspace}
\usepackage{floatrow}
\usepackage{siunitx}
\usepackage{array,multirow,graphicx}
 \usepackage{float}

\newfloatcommand{capbtabbox}{figure}[][\FBwidth]
\usepackage{tikz}

\usepackage{listings}
\usepackage{xcolor}

\definecolor{codegreen}{rgb}{0,0.6,0}
\definecolor{codegray}{rgb}{0.5,0.5,0.5}
\definecolor{codepurple}{rgb}{0.58,0,0.82}
\definecolor{backcolour}{rgb}{0.95,0.95,0.92}

\lstdefinestyle{mystyle}{
    backgroundcolor=\color{backcolour},   
    commentstyle=\color{codegreen},
    keywordstyle=\color{magenta},
    numberstyle=\tiny\color{codegray},
    stringstyle=\color{codepurple},
    basicstyle=\ttfamily\footnotesize,
    breakatwhitespace=false,         
    breaklines=true,                 
    captionpos=b,                    
    keepspaces=true,                 
    numbers=left,                    
    numbersep=5pt,                  
    showspaces=false,                
    showstringspaces=false,
    showtabs=false,                  
    tabsize=2
}

\begin{document}

\date{}

\title{Towards Explainable Evolution Strategies with Large Language Models
}

\author{Jill Baumann and Oliver Kramer
%
% Optional short acknowledgment: remove next line if non-needed
%\thanks{This is an optional funding source acknowledgement.}
%
% DO NOT MODIFY THE FOLLOWING '\vspace' ARGUMENT
\vspace{.3cm}\\
%
% Addresses and institutions (remove "1- " in case of a single institution)
Department of Computing Science\\
Carl von Ossietzky Universität Oldenburg \\
26111 Oldenburg, Germany\\
\url{jill.baumann@uni-oldenburg.de}
%
% Remove the next three lines in case of a single institution
}

\maketitle
\thispagestyle{empty}

\begin{abstract}
This paper introduces an approach that integrates self-adaptive Evolution Strategies (ES) with Large Language Models (LLMs) to enhance the explainability of complex optimization processes. By employing a self-adaptive ES equipped with a restart mechanism, we effectively navigate the challenging landscapes of benchmark functions, capturing detailed logs of the optimization journey. The logs include fitness evolution, step-size adjustments and restart events due to stagnation. An LLM is then utilized to process these logs, generating concise, user-friendly summaries that highlight key aspects such as convergence behavior, optimal fitness achievements, and encounters with local optima. Our case study on the Rastrigin function demonstrates how our approach makes the complexities of ES optimization transparent. Our findings highlight the potential of using LLMs to bridge the gap between advanced optimization algorithms and their interpretability.

\end{abstract}

\section{Introduction}
Nowadays, Large Language Models (LLMs) excel at solving diverse tasks by leveraging the attention mechanism, which enables them to recognize long-range dependencies in texts. Explainable AI (XAI) states an AI model, its impact and potential biases helping to distinguish transparency in AI systems decisions. This approach aims to enhance AI systems' understandability by utilizing an LLM to generate user-friendly explanations of the Evolution Strategies (ES)\cite{es} optimization process, which we refer to as Explainable ES (XES). XES can be used to draw conclusions from the LLM's explanations for potential hyperparameter tuning, such as adjusting the step size or other parameters in our evolutionary strategy use case. Through a detailed case study on the Rastrigin function, we demonstrate the efficacy of our approach in enhancing the understanding of ES, underscoring its potential implications for the field of XAI.

In Section \ref{sec:related}, we explore the landscape of ES and XAI. In Section \ref{sec:xes} we detail our framework XES. Following this, Section~\ref{sec:exp} presents an empirical analysis as a case study using the Rastrigin function. Finally, Section \ref{sec:cons} summarizes our key discoveries and outlines issues for future research.

\section{Related Work}
\label{sec:related}

XAI employs strategies to render AI systems' decisions transparent. Intrinsically interpretable models like decision trees offer direct clarity, while complex models may be simplified for better understanding. Post-hoc algorithms, such as SHAP \cite{shap} and LIME \cite{lime}, provide insights by highlighting influential features or input parts but require white box access and are computationally intensive.
Example-based explanations, including counterfactuals, clarify model decisions by presenting alternative scenarios. Interactive tools and comprehensive documentation further bridge the gap between AI complexity and user comprehension, enhancing transparency and trust. 

Tools like FaAIr, HELM, and ANE demonstrate how LLMs comes to responsible decisions.
Zhao et al. \cite{xai_llm_survey} categorize various explainability techniques for LLMs.
Liu et al. \cite{gpt-medical} introduce a ChatGPT-aided explainable framework for medical image diagnosis combining CLIP and ChatGPT that increases the zero-shot image classification accuracy on five medical datasets. Kroeger et al. \cite{post-hoc} utilize the in-context learning capability of an LLM to explain the predictions of other models. Experiments conducted on real datasets demonstrate that this approach is competitive with state-of-the-art methods. Chuang et al. \cite{llm_faithful} explore the potential of LLMs as explainers by introducing a framework designed to provide faithful explanations, accurately capturing the prediction behaviors of LLMs.

\section{XES: ES with LLM Explanations}
\label{sec:xes}

XES is based on LLM-generated explanations summarizing ES optimization runs. In our XES variant we employ a self-adaptive ES augmented with a restart mechanism to navigate complex optimization landscapes. The algorithm dynamically adjusts its mutation step sizes based on evolutionary history, enhancing exploration and exploitation capabilities. This self-adaptation mechanism fine-tunes the mutation parameters in response to the algorithm's performance, enabling a more efficient search process.

To counteract the potential for stagnation the algorithm contains a strategic restart mechanism. This mechanism activates when the algorithm detects a prolonged lack of fitness improvements, indicating a local optimum. Upon activation, the algorithm reinitializes its state, including the population and mutation parameters, thereby diversifying the search space and increasing the probability of escaping suboptimal regions.

Critical to our approach is the implementation of a detailed logging system that records the progression of the optimization process. The log file captures key metrics such as the development of fitness values across iterations, the evolution of mutation step sizes, stagnation events and subsequent restarts, offering a comprehensive view of the algorithm's dynamics.

An LLM is employed to transform the extensive data recorded in the log-file into concise, user-friendly narratives, describing the optimization process. The LLM is prompted with prompts structured according to four different prompt strategies like Zero-Shot, Few-Shot, Chain-of-Thought (CoT), and Few-Shot CoT Prompting. 
To emphasize critical aspects of the ES's performance, prompts are manually structured according to their prompt strategy, as detailed in Table~\ref{Tab:prompts}. Due to the way an LLM works, it is not guaranteed that the LLM will provide exactly the same response for the same prompt.

\begin{table}[h!]
  \centering
  \begin{tabularx}{\textwidth}{ l X }
    \toprule
    Prompt strategy & Prompt \\
    \midrule
    Zero-Shot& \textit{"Provide a summary of insights derived from the log file below, delimited by triple backticks, representing a minimization problem. Concentrate on best, worst, and mean fitness values, as well as convergence behavior and local optima. 
    Ensure accurate identification of the best and worst fitness values and calculate the mean fitness over all iterations. 
    Remember, in this context, the lowest value denotes the best fitness, while the highest value represents the worst fitness."} 
    \newline \textit{\`{}\`{}\`{}\{log file\}\`{}\`{}\`{}}\\
    \hline
    Few-Shot & \textit{Zero-Shot Prompt \newline \{Correct Answer\} \newline Zero-Shot Prompt}\\
    \hline
    CoT & \textit{"Provide a summary of insights derived from the log file below, delimited by triple backticks, representing a minimization problem.
    Ensure accurate identification of the best and worst fitness values and calculate the mean fitness over all iterations. 
    Remember, in this context, the lowest value denotes the best fitness, while the highest value represents the worst fitness.\newline
    1. Begin by extracting the best fitness value achieved.\newline
    2. Next, extract the worst fitness value achieved during the optimization process.\newline
    3. Calculate the average fitness value across all iterations.\newline
    4. Analyze the convergence behavior, including trends or patterns indicating convergence behavior.\newline
    5. Lastly, determine if there are any instances of reaching local optima or encountering plateaus."} 
    \newline \textit{\`{}\`{}\`{}\{log file\}\`{}\`{}\`{}}\\
    \hline
    Few-Shot CoT & \textit{CoT Prompt \newline \{Correct Answer\} \newline CoT Prompt} \\
    \bottomrule
  \end{tabularx}
  \caption{Prompts according to their prompt strategy.}\label{Tab:prompts}
\end{table}

These instruction guides the LLM to focus on the pivotal elements defining the optimization journey: the development towards convergence, the attainment of optimal fitness values and interactions with local optima, including restarts due to stagnation. This process requires the LLM to view the detailed log data and summarize the most important information in a comprehensible summary.

\section{Case Study}
\label{sec:exp}

In this section we present a case study, which is a practical application of our self-adaptive ES with LLM-generated explanations, utilizing the highly multimodal Rastrigin function to demonstrate the efficacy of our approach.

\subsection{Setting}

In our study, we deployed a $\sigma$-self-adaptive $(\mu + \lambda)$-ES to optimize a 10-di\-men\-sion\-al Rastrigin function, a challenging benchmark known for its complex landscape with numerous local minima that requires the detection of stagnation in local optima and the triggering of restart mechanisms.
The algorithm was set to run until either a maximum of 10,000 iterations were reached or a fitness threshold of \(10^{-5}\) was achieved. To monitor and analyze the optimization process, logs were systematically generated every 30 iterations of the Rastrigin function, detailing the fitness values and step sizes. Additionally, instances of stagnation and subsequent restarts, particularly at the initial point \((1, \ldots, 1)\), were recorded to understand the algorithm's behavior in trapping and escaping local optima. Overall, XES was tested with three log files of varying lengths. Log file 1 ("short") documents 150 iterations, log file 2 ("middle") spans 420 iterations and log file 3 ("long") consists of 1260 iterations which encompass the highest number of restarts. Listing \ref{lst:logfile_1} presents log file 1. 
To translate these extensive logs into comprehensible narratives, we employed four different language models, setting the temperature to 0.0 for more deterministic text generation.
The LLMs we use are Llama2:70B, Llama3:70B, Mistral 7B and Mixtral 8x7B. 
The prompts followed one of the four prompt strategies' structures.
Each combination of LLM and prompt strategy was repeated ten times per log file. To evaluate the LLM's response, numerical information, such as the best and worst fitness, was automatically extracted and verified whether it was correct. Non-numerical information, such as convergence behavior and local optima, was assessed manually. The total score for a combination of LLM and prompt strategy was determined by assigning one point to each correct statement (best fitness, worst fitness, convergence behavior, local optima) and then normalizing the score to a value between 0 and 1.

\begin{lstlisting}[language=Python, caption=Log file 1 ("short"), label=lst:logfile_1]
Iteration 30: Fitness: 1.9899, Step size: 8.5475e-07
Iteration 60: Fitness: 1.9899, Step size: 3.8096e-10
Iteration 90: Fitness: 1.9899, Step size: 5.0931e-10
Iteration 120: Fitness: 1.9899, Step size: 7.7127e-11
Restarting at iteration 149 due to stagnation
Iteration 150: Fitness: 5.0554, Step size: 0.4503

\end{lstlisting}

\subsection{Results}

Table \ref{tab:results_logfile1} shows the results of the experiments conducted w.r.t. the average score achieved across the ten repetitions.

\begin{table}[H]
  \centering
  \caption{Comparison of four different prompt strategies across three log files.}
  \label{tab:results_logfile1}
  \begin{tabular}{@{} l c c c c c @{}}
    \toprule
    & & \multicolumn{1}{c}{Zero-Shot} & \multicolumn{1}{c}{Few-Shot} & \multicolumn{1}{c}{CoT} & \multicolumn{1}{c}{Few-Shot CoT}  \\
    \midrule
    Llama2:70B & \textit{short} & 0.76 & \textbf{1.0} & 0.33 & \textbf{0.98}\\
    & \textit{middle} & 0.45 & 0.55 & 0.55 & 0.35\\
    & \textit{long} & 0.35 & 0.35 & 0.38 & 0.3\\
    \hline
    Llama3:70B & \textit{short} & 0.7 & \textbf{1.0} & 0.53 & 0.95\\
    & \textit{middle} & 0.7 & 0.73 & 0.75 & 0.73\\
    & \textit{long} & 0.85 & \textbf{1.0} & 0.8 & 0.88\\
    \hline
    Mistral 7B & \textit{short} & 0.78 & 0.88 & 0.7 & 0.93\\
    & \textit{middle} & 0.5 & 0.85 & 0.7 & 0.38 \\
    & \textit{long} & 0.45 & 0.55 & 0.5 & 0.58\\
    \hline
    Mixtral 8x7B & \textit{short} & 0.93 & 0.98 & 0.83 & 0.88\\
    & \textit{middle} & 0.8 & \textbf{1.0} & 0.78 & 0.65 \\
    & \textit{long} & \textbf{0.95} & 0.93 & \textbf{0.9} & 0.95\\
    \bottomrule
  \end{tabular}
\end{table}

The outcomes of the conducted experiments are largely similar. Notably, Llama2:70B shows lower scores, whereas Mixtral 8x7B, especially with Few-Shot Prompting, outperforms the others, achieving a peak performance value of 1.0. Among the different prompt strategies, Few-Shot Prompting achieves the best average results across all models.
For instance Mixtral 8x7B with Few-Shot-Prompting for log file 2 outputs:

\textit{"The algorithm initially converged to a local optimum near 3.98, then re\-start\-ed and converged to a lower optimum near 2, followed by another restart and convergence to the best observed value of 0.0106. Multiple local optima were encountered during the optimization process. The lowest fitness value observed was 0.0106. The highest fitness value observed was 3.9852."}

The LLM's summary of the ES run is clear and coherent, detailing key aspects such as best and worst fitness values, and critical events like convergence and stagnation. 
The inclusion of specific numerical details enhances the summary's precision, providing a transparent view of the ES's performance on the Rastrigin function.
However, the analysis could benefit from further context on the sig\-nif\-i\-cance of step size adjustments and a more explicit evaluation of the strategy's overall success. 

XES tends to yield superior results with shorter log files, providing more detailed responses that mention specific iterations and values. In contrast, longer log file responses result in more general information, often referencing iterations and values from the log file's last iterations which is typical for the way an LLM works.
A limitation is the parameter context length of an LLM, which may not be exceeded when formulating prompts based on the prompt strategy and the log file's content. Developing techniques to shorten long log files could enhance robustness. 
The best and worst fitness values can also be extracted deterministically. This approach demonstrates the LLM's capability to extract these values, making it an appropriate alternative for obtaining values and providing user-friendly explanations of a optimization process in an easy manner.

\section{Conclusions}
\label{sec:cons}

Our approach leverages the advanced natural language generation capabilities of LLMs to transform technical optimization logs into accessible explanations, thereby providing the intricacies of the ES optimization process for a broader audience. Looking ahead, we see substantial potential in further enriching this framework. Firstly, by integrating an interactive analysis layer that prompts user inquiries, so we can tailor explanations to individual needs, enhancing user comprehension. Secondly, the insights derived from LLM analyses could be used in agent-based systems capable of automating optimization actions.

\bibliographystyle{abbrv}
\bibliography{literature}

\end{document}